\documentclass[conference]{IEEEtran}
\IEEEoverridecommandlockouts
\usepackage{cite}
\usepackage{amsmath,amssymb,amsfonts}
\usepackage{algorithmic}
\usepackage{graphicx}
\usepackage{textcomp}
\usepackage{xcolor}
\usepackage{booktabs}
\usepackage{multirow}
\def\BibTeX{{\rm B\kern-.05em{\sc i\kern-.025em b}\kern-.08em
    T\kern-.1667em\lower.7ex\hbox{E}\kern-.125emX}}
\begin{document}

\title{TFGformer: Multivariate Time Series Forecasting via Time-Frequency Graph Learning and Covariate Fusion\\

\thanks{This work is supported by the National Natural Science Foundation of China under Grants 62272052}
}

\author{\IEEEauthorblockN{Yu Sun}
\IEEEauthorblockA{
\textit{Beijing University of}\\
\textit{Posts and Telecommunications}\\
Beijing, China \\
sunyu7@bupt.edu.cn}
\and
\IEEEauthorblockN{Yuan Chang}
\IEEEauthorblockA{\textit{China Telecom } \\
\textit{Research Institute}\\
Beijing, China \\
changy8@chinatelecom.cn
}
\and
\IEEEauthorblockN{Xiaohou SHI}
\IEEEauthorblockA{\textit{China Telecom } \\
\textit{Research Institute}\\
Beijing, China \\
shixh6@chinatelecom.cn
}
\and
\IEEEauthorblockN{ Yan Sun}
\IEEEauthorblockA{
\textit{Beijing University of}\\
\textit{Posts and Telecommunications}\\
Beijing, China \\
sunyan@bupt.edu.cn}
}
\maketitle

\begin{abstract}
Multivariate time series forecasting requires modeling complex temporal dynamics and inter-variable dependencies. However, in practice, noisy correlations often introduce interference, while crucial contextual covariates such as calendar features and external events are frequently underutilized. This highlights the need for models that selectively identify meaningful relationships while effectively incorporating auxiliary information. To address these challenges, we propose a unified framework integrating time–frequency graph structure learning with covariate-aware representation fusion. The Time-Frequency Graph (TFG) module leverages the Short-Time Fourier Transform (STFT) to capture complementary temporal–spectral patterns. It then learns adaptive inter-variable correlations via a weighted Mahalanobis distance and employs Gumbel–Softmax sampling to derive sparse, dynamic graphs that suppress irrelevant dependencies. In parallel, MLP-based modules fuse historical and future covariates into the main sequence representation, enabling the model to exploit trend priors and contextual signals.Extensive experiments on electricity, traffic, and meteorology benchmarks demonstrate that our approach consistently outperforms state-of-the-art Transformer-based models, confirming its effectiveness in modeling selective variable interactions and leveraging covariates for improved forecasting accuracy.
\end{abstract}


\section{Introduction}
% With the rapid advancement of deep learning, Transformer-based large-scale models have achieved remarkable success in natural language processing \cite{b1} and computer vision\cite{b2}, and have simultaneously spurred the development of foundation models for multivariate time series forecasting (MTSF). Time series forecasting aims to uncover underlying dynamic patterns and trends from historical observations and plays a crucial role in finance, transportation, meteorology, and network operations.

% As massive and increasingly complex datasets become more prevalent, the dependencies among variables in multivariate time series have grown more intricate and dynamic. Moreover, long-term forecasting is particularly susceptible to error accumulation and performance degradation\cite{b0}. Consequently, effectively capturing these complex, time-varying inter-variable relationships while fully leveraging external covariates enriched with prior information remains a key challenge in contemporary MTSF research.  
Transformer-based models have achieved remarkable success in NLP [1] and vision [2], spurring the development of foundation models for multivariate time series forecasting (MTSF). MTSF uncovers dynamic patterns from historical data, playing a crucial role in various real-world applications. As datasets grow complex, inter-variable dependencies become more intricate. Since long-term forecasting naturally suffers from error accumulation [3], effectively capturing these time-varying relationships while leveraging external covariates remains a key challenge.

Existing approaches attempt to address inter-variable dependencies in various ways. For example, PatchTST \cite{b3} segments sequences into patches and employs channel-independent processing to mitigate interference from weakly correlated variables; however, it lacks explicit cross-variable modeling and interpretability. iTransformer \cite{b4} introduces channel-mixing self-attention to capture variable-level dependencies, yet it overlooks fine-grained temporal dynamics within individual variables. TimerXL \cite{b5} improves upon this by jointly modeling intra- and inter-variable patterns, but it still assumes static associations across all variable pairs, failing to account for the strength and temporal evolution of these dependencies. This limitation becomes particularly problematic in scenarios where relationships are sparse or change over time.

Meanwhile, covariates such as calendar attributes, device metadata, and event-driven signals carry valuable prior knowledge that can significantly enhance forecasting accuracy. Yet, most existing models incorporate them through simple concatenation or by treating them as extra channels, which often introduces noise and overlooks their inherent structure. Although models like TiDE \cite{b6} and ChronosX \cite{b7} highlight the benefits of dedicated covariate fusion, TiDE is primarily an MLP-based architecture, while ChronosX is tailored for univariate, token-based forecasting. Consequently, neither is ideally suited for explicitly modeling dynamic inter-variable dependencies in a multivariate, patch-based setting.

To address these limitations, this paper proposes a novel multivariate time series forecasting framework that integrates time–frequency graph learning with effective covariate fusion. Our main contributions are threefold:
\begin{itemize}
    \item We design a novel time-frequency graph structure learning module that leverages Short-Time Fourier Transform (STFT) and Mahalanobis distance to capture dynamic and nonlinear dependencies among variables. By estimating pairwise relational probabilities in the joint time–frequency domain, the module adaptively constructs an attention mask that guides the attention mechanism, enabling more accurate modeling of complex inter-variable interactions.
    \item We introduce an MLP-based covariate fusion module that integrates both historical and future covariates with the primary sequence representation through residual connections, without losing the original data characteristics. This design allows the model to fully exploit prior trend knowledge and external dynamic context.
    \item We conduct comprehensive experiments on seven widely used benchmark datasets spanning the electricity, traffic, and weather domains. The results demonstrate that our approach consistently achieves state-of-the-art performance across multiple forecasting horizons and evaluation metrics, validating its effectiveness in modeling dynamic inter-variable relationships and leveraging covariate information.
\end{itemize}

\section{Related Works}

\subsection{Transformer-Based Time Series Forecasting}

Transformers show strong potential in MTSF. Early works like Informer and Autoformer [8] improved attention for long-term dependencies. However, effectively modeling complex inter-variable relationships remains a significant challenge.

Existing Transformer-based models for multivariate time series generally fall into two categories.
Channel-independent \cite{b9} approaches, such as PatchTST, model each variable separately, focusing on intra-variable temporal patterns. Although effective, they do not explicitly capture interactions between variables.
Channel-mixing \cite{b10} approaches like iTransformer treat each variable as a token and apply attention across the variable dimension, thereby modeling inter-variable relationships but potentially overlooking fine-grained temporal dynamics within each variable.

To leverage the strengths of both directions, recent hybrid designs such as TimerXL combine patching strategies with channel-mixing mechanisms, enabling simultaneous modeling of intra- and inter-variable patterns. Nevertheless, these models implicitly assume that all variable pairs share equal levels of correlation. They are unable to distinguish the varying strengths or temporal evolution of relationships among variables, which may introduce noise when dependencies are sparse or highly dynamic.

\subsection{Frequency-Domain Analysis in Time Series Modeling}

Recent research has revealed that frequency-domain features offer natural advantages in depicting multivariate synchrony and periodic consistency. Motivated by this, several methods have shifted from pure time-domain modeling to frequency-domain representations \cite{b11}, leveraging amplitude and phase information \cite{b12} to characterize inter-variable relationships. Nevertheless, most existing approaches rely on global Fourier transforms, which map the entire sequence into holistic frequency components, thereby overlooking temporal locality. As a result, they struggle to capture the non-stationary and time-varying correlation structures that are common in real-world multivariate time series.

\subsection{Covariate Utilization in Time Series Forecasting}

In practical forecasting tasks, auxiliary covariates such as weather conditions, calendar features, and holiday indicators provide valuable prior information. Despite this, mainstream forecasting models typically adopt relatively basic fusion strategies. One common approach encodes covariates in a simple manner and then concatenates them with the main series. Another approach directly treats covariates as additional variables and processes them together with the primary inputs. However, because heterogeneous covariates exhibit distinct temporal patterns and statistical properties, such naive concatenation or joint modeling often introduces noise and prevents the model from effectively extracting useful covariate information.As an attempt to improve this, models like ChronosX have introduced a fusion module. However, it is noteworthy that its design is targeted at univariate scenarios and operates on a token-based framework, rather than patch-based.

\section{Methodology}

\subsection{Problem Definition}\label{AA}
This study focuses on the task of multivariate time series forecasting.
Given a historical observation sequence ${X} = \{x_1, x_2, \ldots, x_L\} \in \mathbb{R}^{L \times C}$, where $L$ is the sequence length and $C$ denotes the number of variables.We additionally consider historical covariates ${C}_{\text{history}} \in \mathbb{R}^{L \times C_z}$,and future covariates ${C}_{\text{future}} \in \mathbb{R}^{S \times C_z}$, which provide auxiliary information such as calendar features, weather indicators, or other exogenous factors.

The forecasting model aims to generate future series values based on the historical observations and covariates:
$\hat{{Y}} = f({X}, {C}_{\text{his}}, {C}_{\text{fut}})$, where $\hat{{Y}} \in \mathbb{R}^{S \times C}$ denotes the predicted multivariate sequence for the next $S$ time steps.
\subsection{Structure Overview}

\begin{figure*}[htbp]
\centering
\includegraphics[width=0.96\linewidth]{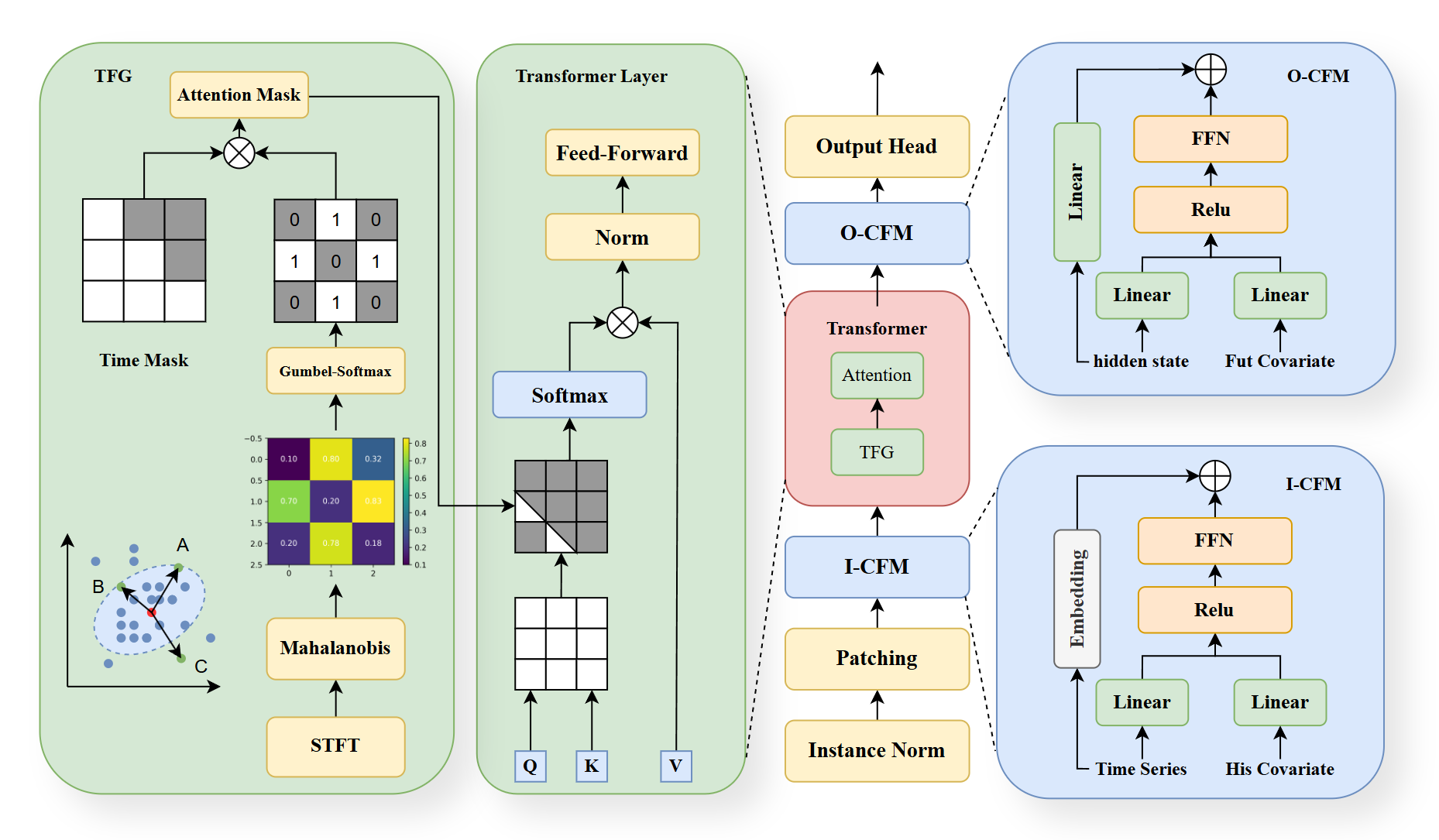}
\caption{The overall architecture of TFGformer.}
\label{fig}
\end{figure*}

Our model architecture is illustrated in Fig.~\ref{fig} The proposed model consists of three main components: the Time–Frequency Graph (TFG) module, the Covariate Feature Fusion (CFM) module, and a Transformer-based sequence modeling module.

Given a multivariate time series along with historical and future covariates, the TFG module first analyzes temporal and spectral patterns to learn dynamic inter-variable relationships, producing an attention mask that guides cross-variable interactions. The CFM module is divided into two sub-modules: the Input Covariate Fusion (I-CFM) integrates historical covariates with the original time series before entering the Transformer, while the Output Covariate Fusion (O-CFM) incorporates future covariates after Transformer processing to further refine the hidden representations.

The enriched sequence, containing relational information from TFG and covariate-aware features from I-CFM, is fed into the Transformer backbone. The attention mask generated by the TFG module is applied during attention computation to explicitly model relevant inter-variable dependencies. After Transformer processing, future covariates are fused via O-CFM, and the final forecasting head produces the multivariate predictions for the target horizon.
\subsection{Basic Components}
\begin{itemize}
    \item \textbf{Normalization} Per-channel Z-score \cite{b13} normalization is applied along the temporal dimension:
    \begin{equation}
    \hat{x}_{i,t} = \frac{x_{i,t} - \mu_i}{\sigma_i}
    \label{eq:normal}
    \end{equation}

    \item \textbf{Patching} The input sequence is split into fixed-length temporal patches using a sliding window:
    \begin{equation}
    N = \frac{L - l}{s} + 1
    \label{eq:patch}
    \end{equation}

    \item \textbf{Output Head} A linear layer maps the Transformer outputs to the prediction sequence:
    \begin{equation}
    {\hat{Y}} = {H} {W} + {b}
    \end{equation}
\end{itemize}
\subsection{Time–Frequency Graph Learning Module}
For each variable $\mathnormal{x_c}$ in the multivariate sequence, the module first computes its short-time Fourier transform (STFT) to obtain the magnitude spectrogram. Where $\mathnormal{n_\mathnormal{fft}}$ denotes the FFT size that determines the number of frequency bins $F = \frac{n_{\mathrm{fft}}}{2} + 1$, $\mathnormal{hop}$ is the hop length controlling the temporal stride of successive frames, and $\mathnormal{win}$ specifies the window length used in the analysis. Subsequently, the time-frequency dimensions are flattened to form the channel feature vectors $Z$. where $N_{feat} = F \cdot T$ is the dimensionality of the flattened time-frequency feature space.
\begin{equation} M_c = \left| \mathrm{STFT}(x_c, n_{\mathrm{fft}}, \mathrm{hop}, \mathrm{win}) \right| \in \mathbb{R}^{F \times T} \end{equation} 
\begin{equation} Z = \mathrm{Flatten}(M_c) \in \mathbb{R} ^ {B \times C \times N_{feat}} 
\end{equation}

To adaptively discover the interrelationships among channels, we adopt a learnable Mahalanobis distance \cite{b14} metric to evaluate the pairwise distances over the feature vectors ${Z}$. This metric moves beyond simple weighted distances by learning a matrix that captures the covariance structure of the time-frequency feature space. The distance ${D}_{i, j}$ between the feature vectors of channel $i$ (${Z}_i$) and channel $j$ (${Z}_j$) is defined as a quadratic form:
\begin{equation} {D}_{i, j} = ({Z}_i - {Z}_j)^T {Q} ({Z}_i - {Z}_j) 
\end{equation}
where ${D}\in\mathbb{R}^{B \times C \times C}$ is the resulting distance matrix, and ${Q} \in \mathbb{R}^{N_{feat} \times N_{feat}}$ is a learnable symmetric positive semi-definite (PSD) matrix.We then convert the distance into a similarity (or unnormalized probability) using the inverse relationship and mask out the diagonal elements (distance from self to self):
\begin{equation}
\tilde{{P}}_{i, j} = \begin{cases} \frac{1}{{D}_{i, j} + \epsilon} & \text{if } i \neq j \\ 0 & \text{if } i = j \end{cases}
\end{equation}

To obtain the final probabilistic matrix ${P} \in [0, 1]^{B \times C \times C}$, we perform a normalization step. We utilize max normalization on the non-diagonal elements to stabilize the gradients and preserve relative distances, followed by re-inserting the diagonal elements. 
\begin{equation}
{P} = \frac{\tilde{{P}}}{\max(\tilde{{P}}, \text{dim}=-1) + \epsilon}  + {I}
\end{equation}

To transform the continuous probabilistic matrix ${P}$ into a discrete binary adjacency matrix suitable for downstream attention operations, we treat each element ${P}_{ij}$ as the success probability of a Bernoulli random variable. Specifically, we sample a binary matrix ${G} \in [0, 1]^{B \times C \times C}$ according to
\begin{equation}
G_{ij} \sim \mathrm{Bernoulli}(P_{ij})
\end{equation}
where a higher probability ${P}_{ij}$ corresponds to a greater likelihood that $G_{ij} = 1$, indicating a stronger inferred dependency between channel $\mathnormal{i}$ and channel $\mathnormal{j}$. To allow gradient propagation through this sampling process, we employ the Gumbel–Softmax \cite{b15} reparameterization trick.
\begin{equation}
G = \mathrm{GumbelSoftmax} ({G}_{ij})
\end{equation}

The resulting binary mask ${G}$ effectively filters out noise and weak correlations, ensuring that subsequent computations focus only on the most relevant inter-channel paths learned from the time-frequency domain. To integrate this learned spatial dependency with the temporal dynamics, we first define a causal time mask ${M}_{\text{time}} \in \{0, 1\}^{N \times N}$ as a lower triangular matrix. The final Attention Mask ${M}_{\text{attention}}$ is then constructed by applying the Kronecker \cite{b16} Product ($\otimes$) between the expanded channel mask ${G}$ and the time mask ${M}_{\text{time}}$:
\begin{equation}
{M}_{\text{attention}} = {G} \otimes {M}_{\text{time}}
\end{equation}

The computational complexity of the TFG module scales efficiently with the sequence length $L$ and the variable count $C$. First, the frequency feature extraction requires $\mathcal{O}(C \cdot L \log L)$ time. Second, for graph structure learning, we utilize a learnable diagonal weight vector optimized via tensor broadcasting. This specific design reduces the relation computation complexity to $\mathcal{O}(C^2 \cdot L)$. 

\subsection{Covariate Fusion Module}

To effectively incorporate external signals into the forecasting pipeline, the model introduces a covariate fusion module consisting of two components: the Input Covariate Fusion Module (I-CFM) and the Output Covariate Fusion Module (O-CFM). Both modules rely on MLP-based transformations to fuse covariates with the model's internal representations, enabling the network to adjust its learned features according to historical and future contextual factors.

I-CFM operates at the input stage and injects past covariate information into the patch embeddings. Let ${X}_{\text{emb}}$ denote the patch embedding obtained from the original input sequence $X$. This embedding ${X}_{\text{emb}}$ and the projected past covariates ${C}_\mathnormal{his}$ are concatenated, processed through a ReLU activation, and then passed through an MLP-based feedforward network. The output is added to the original embedding ${X}_{\text{emb}}$ through a residual path to obtain a covariate-aware input representation:
\begin{equation}
    Z_{\text{in}} = X_{\text{emb}} + \mathrm{FFN}\!\left(
    \mathrm{ReLU}\!\left(
        [\, W_x X \,;\, W_c C_{\text{his}} \,]
    \right)
\right)
\end{equation}

O-CFM operates at the decoder side and fuses future covariates with the Transformer-generated hidden states. The hidden state ${H}$ is first projected through a linear layer, concatenated with projected future covariates ${C}_\mathnormal{fut}$, and passed through a ReLU activation followed by the MLP. The fused output is added to the projected hidden state to modulate the decoder representation:
\begin{equation}
Z_{\text{out}} = W_h H + \mathrm{FFN}\!\left(
    \mathrm{ReLU}\!\left(
        [\, W_h H \,;\, W_f C_{\text{fut}} \,]
    \right)
\right)
\end{equation}

In practical forecasting, future covariates $C_{future}$ may contain missing values. To ensure the robustness of the O-CFM, we introduce a preprocessing step: missing continuous covariates are forward-filled, while categorical ones are zero-padded. This ensures the MLP-based fusion remains stable without propagating noise into the hidden states.

\subsection{Transformer Backbone}
The Transformer architecture is employed to process the covariate-enhanced input patches $Z_{\text{in}} \in \mathbb{R}^{B \times (C \cdot N) \times D}$. Multi-head self-attention is employed. To enforce the relevant channel interactions learned by the TFG module and to ensure causality, we apply the attention mask additively before the softmax operation.
\begin{equation}
{{Scores}} =  \frac{{Q}\cdot {K}^\top}{\sqrt{d_k}} \odot {M}_{} + (1- M) \odot (- \infty)
\end{equation}
\begin{equation}
Out = Softmax(Scores) \cdot V
\end{equation}
\section{Experiment}
\subsection{Experimental Settings}
% \begin{table}[b]
%     \centering
%     \caption{DATASET DESCRIPTIONS.}
%     \label{tab:dataset_descriptions}
%     \setlength{\tabcolsep}{8pt} % 可选：调整列间距以匹配图片宽度
%     \begin{tabular}{cccc}
%         \toprule
%         Dataset & Number of Series & Timestamps & Temporal Granularity \\
%         \midrule
%         Electricity & 321 & 26304 & 1 hour \\
%         \midrule
%         Weather & 21 & 52696 & 10 mins \\
%         \midrule
%         ETTh & 7 & 17420 & 1 hour \\
%         \midrule
%         ETTm & 7 & 69680 & 15 mins \\
%         \midrule
%         Traffic & 862 & 17544 & 1 hour \\
%         \bottomrule
%     \end{tabular}
% \end{table}
\begin{table*}[!t]
    \centering
    \caption{MULTIVARIATE LONG-TERM FORECASTING RESULTS.}
    \label{tab:main_results}
    \small 
    \setlength{\tabcolsep}{4pt} 
    \begin{tabular}{@{}c|c|cc|cc|cc|cc|cc|cc|cc@{}}
        \toprule
        \multirow{2}{*}{\textbf{Dataset}} & \multirow{2}{*}{\textbf{Horizon}} & 
        \multicolumn{2}{c}{\textbf{TFGformer}} &
        \multicolumn{2}{c}{iTransformer} &
        \multicolumn{2}{c}{TiDE} &
        \multicolumn{2}{c}{PatchTST} &
        \multicolumn{2}{c}{Autoformer} &
        \multicolumn{2}{c}{FEDformer} &
        \multicolumn{2}{c}{DLinear} \\
        \cmidrule(lr){3-4} \cmidrule(lr){5-6} \cmidrule(lr){7-8} \cmidrule(lr){9-10} \cmidrule(lr){11-12} \cmidrule(lr){13-14} \cmidrule(lr){15-16}
        & & MSE & MAE & MSE & MAE & MSE & MAE & MSE & MAE & MSE & MAE & MSE & MAE & MSE & MAE \\
        \midrule
        % --- ETTh1 数据集 ---
        \multirow{5}{*}{\rotatebox{90}{ETTh1}}
        & 96  & \underline{0.380} & \textbf{0.399} & 0.386 & 0.405 & 0.383 & 0.403 & 0.414 & 0.419 & 0.449 & 0.459 & \textbf{0.376} & 0.419 & 0.386 & \underline{0.400} \\
        & 192 & \underline{0.432} & 0.437 & 0.441 & 0.436 & 0.437 & \underline{0.433} & 0.460 & 0.445 & 0.500 & 0.482 & \textbf{0.420} & 0.448 & 0.437 & \textbf{0.432} \\
        & 336 & 0.474 & \underline{0.455} & 0.487 & 0.458 & \textbf{0.455} & \textbf{0.453} & 0.501 & 0.466 & 0.521 & 0.496 & \underline{0.459} & 0.465 & 0.481 & 0.459 \\
        & 720 & \textbf{0.485} & \textbf{0.474} & 0.503 & 0.491 & \underline{0.490} & 0.489 & 0.500 & \underline{0.488} & 0.514 & 0.512 & 0.506 & 0.507 & 0.519 & 0.516 \\
        \cmidrule(lr){2-16} 
        & Avg & 0.442 & \textbf{0.441} & 0.454 & 0.447 & \underline{0.441} & \underline{0.445} & 0.469 & 0.454 & 0.496 & 0.487 & \textbf{0.440} & 0.460 & 0.456 & 0.452 \\
        \midrule
        % --- ETTh2 数据集 ---
        \multirow{5}{*}{\rotatebox{90}{ETTh2}}
        & 96  & \textbf{0.294} & \textbf{0.346} & \underline{0.297} & 0.349 & 0.301 & 0.350 & 0.302 & \underline{0.348} & 0.346 & 0.388 & 0.358 & 0.397 & 0.333 & 0.387 \\
        & 192 & \textbf{0.378} & \textbf{0.392} & \underline{0.380} & \underline{0.400} & 0.384 & 0.402 & 0.388 & 0.400 & 0.456 & 0.452 & 0.429 & 0.439 & 0.477 & 0.476 \\
        & 336 & \textbf{0.399} & \textbf{0.430}& 0.428 & \underline{0.432} & \underline{0.420} & 0.434 & 0.426 & 0.433 & 0.482 & 0.486 & 0.496 & 0.487 & 0.594 & 0.541 \\
        & 720 & \underline{0.430} & \underline{0.446} & \textbf{0.427} & \textbf{0.445} & 0.433 & 0.450 & 0.431 & \underline{0.446} & 0.515 & 0.511 & 0.463 & 0.474 & 0.831 & 0.657 \\
        \cmidrule(lr){2-16}
        & Avg & \textbf{0.375}& \textbf{0.403} & \underline{0.383} & \underline{0.407} & 0.385 &0.409 & 0.387 & \underline{0.407} & 0.450 & 0.459 & 0.437 & 0.449 & 0.559 & 0.515 \\
        \midrule
        % --- ETTm1 数据集 ---
        \multirow{5}{*}{\rotatebox{90}{ETTm1}}
        & 96  & \textbf{0.321} & \textbf{0.354} & 0.334 & \underline{0.368} & 0.332 & 0.370 & \underline{0.329} & 0.367 & 0.505 & 0.475 & 0.379 & 0.419 & 0.345 & 0.372 \\
        & 192 & \textbf{0.364} & \textbf{0.380} & 0.377 & 0.391 & 0.380 & 0.401 & \underline{0.367} & \underline{0.385} & 0.553 & 0.496 & 0.426 & 0.441 & 0.380 & 0.389 \\
        & 336 & \textbf{0.384} & \textbf{0.401} & 0.426 & 0.420 & \underline{0.398} & \underline{0.408} & 0.399 & 0.410 & 0.621 & 0.537 & 0.445 & 0.459 & 0.413 & 0.413 \\
        & 720 & \textbf{0.442} & \textbf{0.437} & 0.491 & 0.459 & \underline{0.447} & \textbf{0.437} & 0.454 & \underline{0.439} & 0.671 & 0.561 & 0.543 & 0.490 & 0.474 & 0.453 \\
        \cmidrule(lr){2-16}
        & Avg & \textbf{0.377} & \textbf{0.393} & 0.407 & 0.410 & 0.389 & 0.404 & \underline{0.387} & \underline{0.400} & 0.588 & 0.517 & 0.448 & 0.452 & 0.403 & 0.407 \\ 
        \midrule
        % --- ETTm2 数据集 ---
        \multirow{5}{*}{\rotatebox{90}{ETTm2}}
        & 96  & \textbf{0.170} & \textbf{0.247} & 0.180 & 0.264 & \underline{0.173} & \underline{0.256} & 0.175 & 0.259 & 0.255 & 0.339 & 0.203 & 0.287 & 0.193 & 0.292 \\
        & 192 & \textbf{0.240} & \textbf{0.297} & 0.250 & 0.309 & 0.245 & 0.310 & \underline{0.241} & \underline{0.302} & 0.281 & 0.340 & 0.269 & 0.328 & 0.284 & 0.362 \\
        & 336 & \textbf{0.298} & \textbf{0.337} & 0.311 & 0.348 & \underline{0.304} & \underline{0.341} & 0.305 & 0.343 & 0.339 & 0.372 & 0.325 & 0.366 & 0.369 & 0.427 \\
        & 720 & \textbf{0.394} & \textbf{0.393} & 0.412 & 0.407 & 0.406 & 0.398 & \underline{0.402} & \underline{0.400} & 0.433 & 0.432 & 0.421 & 0.415 & 0.554 & 0.522 \\
        \cmidrule(lr){2-16}
        & Avg & \textbf{0.275} & \textbf{0.318} & 0.288 & 0.332 & 0.282 & \underline{0.326} & \underline{0.281} & \underline{0.326} & 0.327 & 0.371 & 0.305 & 0.349 & 0.350 & 0.401 \\
        \midrule
        % --- Weather 数据集 ---
        \multirow{5}{*}{\rotatebox{90}{Weather}}
        & 96  & \textbf{0.162} & \textbf{0.209} & \underline{0.174} & \underline{0.214} & 0.185 & 0.226 & 0.177 & 0.218 & 0.266 & 0.336 & 0.217 & 0.296 & 0.196 & 0.255 \\
        & 192 & \textbf{0.218} & \textbf{0.249} & \underline{0.221} & \underline{0.254} & 0.237 & 0.265 & 0.225 & 0.259 & 0.307 & 0.367 & 0.276 & 0.336 & 0.237 & 0.296 \\
        & 336 & \textbf{0.271} & \textbf{0.288} & \underline{0.278} & \underline{0.296} & 0.282 & 0.305 & \underline{0.278} & 0.297 & 0.359 & 0.395 & 0.339 & 0.380 & 0.283 & 0.335 \\
        & 720 & \textbf{0.348} & \textbf{0.347} & 0.358 & \textbf{0.347} & 0.359 & 0.353 & \underline{0.354} & \underline{0.348}& 0.419 & 0.428 & 0.403 & 0.428 & 0.345 & 0.381 \\
        \cmidrule(lr){2-16}
        & Avg & \textbf{0.249} & \textbf{0.273} & \underline{0.258} & \underline{0.278} & 0.266 & 0.287 & 0.259 & 0.281 & 0.338 & 0.382 & 0.309 & 0.360 & 0.265 & 0.317 \\
        \midrule
        % --- Electricity 数据集 ---
        \multirow{5}{*}{\rotatebox{90}{Electricity}}
        & 96  & \textbf{0.139} & \textbf{0.227} & \underline{0.148} & \underline{0.240} & 0.174 & 0.271 & 0.181 & 0.270 & 0.201 & 0.317 & 0.193 & 0.308 & 0.197 & 0.282 \\
        & 192 & \textbf{0.160} & \textbf{0.249} & \underline{0.162} & \underline{0.253} & 0.186 & 0.280 & 0.188 & 0.274 & 0.222 & 0.334 & 0.201 & 0.315 & 0.196 & 0.285 \\
        & 336 & \textbf{0.171} & \textbf{0.263} & \underline{0.178} & \underline{0.269} & 0.207 & 0.298 & 0.204 & 0.293 & 0.231 & 0.338 & 0.214 & 0.329 & 0.209 & 0.301 \\
        & 720 & \textbf{0.213} & \textbf{0.298} & \underline{0.225} & \underline{0.317} & 0.249 & 0.329 & 0.246 & 0.324 & 0.254 & 0.361 & 0.246 & 0.355 & 0.245 & 0.333 \\
        \cmidrule(lr){2-16}
        & Avg & \textbf{0.171} & \textbf{0.259} & \underline{0.178} & \underline{0.270} & 0.204 & 0.295 & 0.205 & 0.290 & 0.227 & 0.338 & 0.214 & 0.327 & 0.212 & 0.300 \\
        \midrule
        % --- Traffic 数据集 ---
        \multirow{5}{*}{\rotatebox{90}{Traffic}}
        & 96  & \textbf{0.390} & \textbf{0.259} & \underline{0.395} & \underline{0.268} & 0.450 & 0.289 & 0.462 & 0.295 & 0.613 & 0.388 & 0.587 & 0.366 & 0.650 & 0.396 \\
        & 192 & \textbf{0.412} & \textbf{0.273} & \underline{0.417} & \underline{0.276} & 0.458 & 0.292 & 0.466 & 0.296 & 0.616 & 0.382 & 0.604 & 0.373 & 0.598 & 0.370 \\
        & 336 & \textbf{0.424} & \textbf{0.277} & \underline{0.433} & \underline{0.283} & 0.478 & 0.301 & 0.482 & 0.304 & 0.622 & 0.377 & 0.621 & 0.383 & 0.605 & 0.373 \\
        & 720 & \textbf{0.462} & \textbf{0.297} & \underline{0.467} & \underline{0.302} & 0.497 & 0.313 & 0.514 & 0.322 & 0.660 & 0.408 & 0.626 & 0.382 & 0.645 & 0.394 \\
        \cmidrule(lr){2-16}
        & Avg & \textbf{0.422} & \textbf{0.276} & \underline{0.428} & \underline{0.282} & 0.471 & 0.299 & 0.481 & 0.304 & 0.628 & 0.389 & 0.610 & 0.376 & 0.625 & 0.383 \\
        \midrule
    \end{tabular}
\end{table*}
We conduct extensive experiments on four real-world multivariate time series datasets: ETT (comprising four subsets ETTh1, ETTh2, ETTm1, ETTm2), ECL (321 clients), Traffic (862 road sensors), and Weather (21 meteorological indicators).
Following established practice, we split the data into training, validation, and test sets with a ratio of 6:2:2 for the ETT datasets and 7:1:2 for the remaining datasets. Evaluation protocol. We fix the input length to 96 and evaluate prediction lengths $L \in \{96, 192, 336, 720\}$. Mean Squared Error (MSE) and Mean Absolute Error (MAE) are used as core evaluation metrics. We train our model using the Adam optimizer with a learning rate tuned from $\{1e-4, 5e-4, 1e-3\}$ for each dataset, combined with learning rate annealing and early stopping based on validation performance. The checkpoint achieving the lowest validation error is used for final testing. All experiments are conducted on a single NVIDIA A800-SXM4-40GB GPU. Baseline setting. We select six representative forecasting models as baselines: iTransformer, TiDE, PatchTST, Autoformer , FEDformer, and DLinear .

\subsection{Main Results and Analysis}
The comprehensive multivariate long-term forecasting results are presented in Table \ref{tab:main_results}. The best results are highlighted in bold, and the second-best results are underlined.It is evident that our TFGformer demonstrates outstanding predictive performance, achieving state-of-the-art (SOTA) results by consistently outperforming all baselines. TFGformer achieves the best average MSE and MAE on 6 out of 7 datasets (ETTh2, ETTm1, ETTm2, Weather, Electricity, and Traffic), and remains highly competitive on the final ETTh1 dataset.Compared to recent strong baselines, TFGformer shows significant improvements. On average across all seven datasets, TFGformer reduces MSE by 3.6\% compared to iTransformer, 6.4\% compared to PatchTST, 5.2\% compared to TiDE, and 16.4\% compared to the Transformer-based FEDformer. This indicates that TFGformer successfully enhances the long-term forecasting ability for complex multivariate time series (MTS) prediction problems.

\subsection{Ablation Study}

\begin{figure}[htbp]
\centerline{\includegraphics[width=\columnwidth]{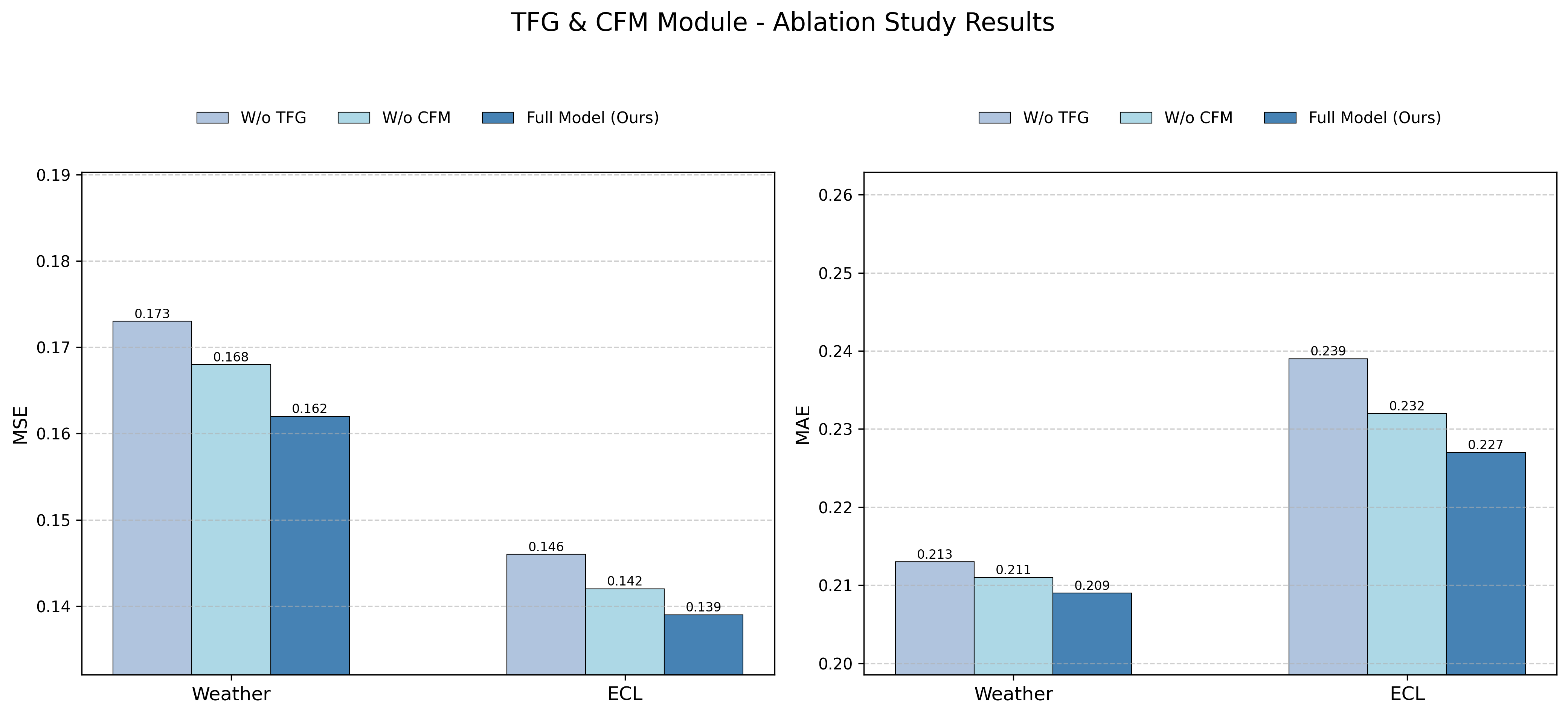}}
\caption{The ablation study results of TFGformer}
\label{fig2}
\end{figure}
To validate the effectiveness of our model's key components, specifically the TFG and CFM modules, we conducted a comprehensive ablation study. We compare our Full Model (Ours) against two variants: (i) W/o TFG, which removes the TFG module, and (ii) W/o CFM, which removes the CFM module.The results on the Weather and ECL datasets are presented in Fig.\ref{fig2} The experimental results clearly demonstrate the contribution of each component. Our Full Model consistently achieves the best performance (lowest MSE and MAE) across all experiments.Quantitatively, removing the CFM module (W/o CFM) resulted in an average MSE increase of 3.0\% compared to the full model across the two datasets. The impact of the TFG module was even more significant; removing it (W/o TFG) caused a larger performance degradation, with an average MSE increase of 6.0\%. These results confirm that both modules are integral to our model's architecture, and each part contributes effectively to enhancing the forecasting performance.

\subsection{Visualization of Variable Correlations}
\begin{figure}[htbp]
\centerline{\includegraphics[width=0.8\columnwidth]{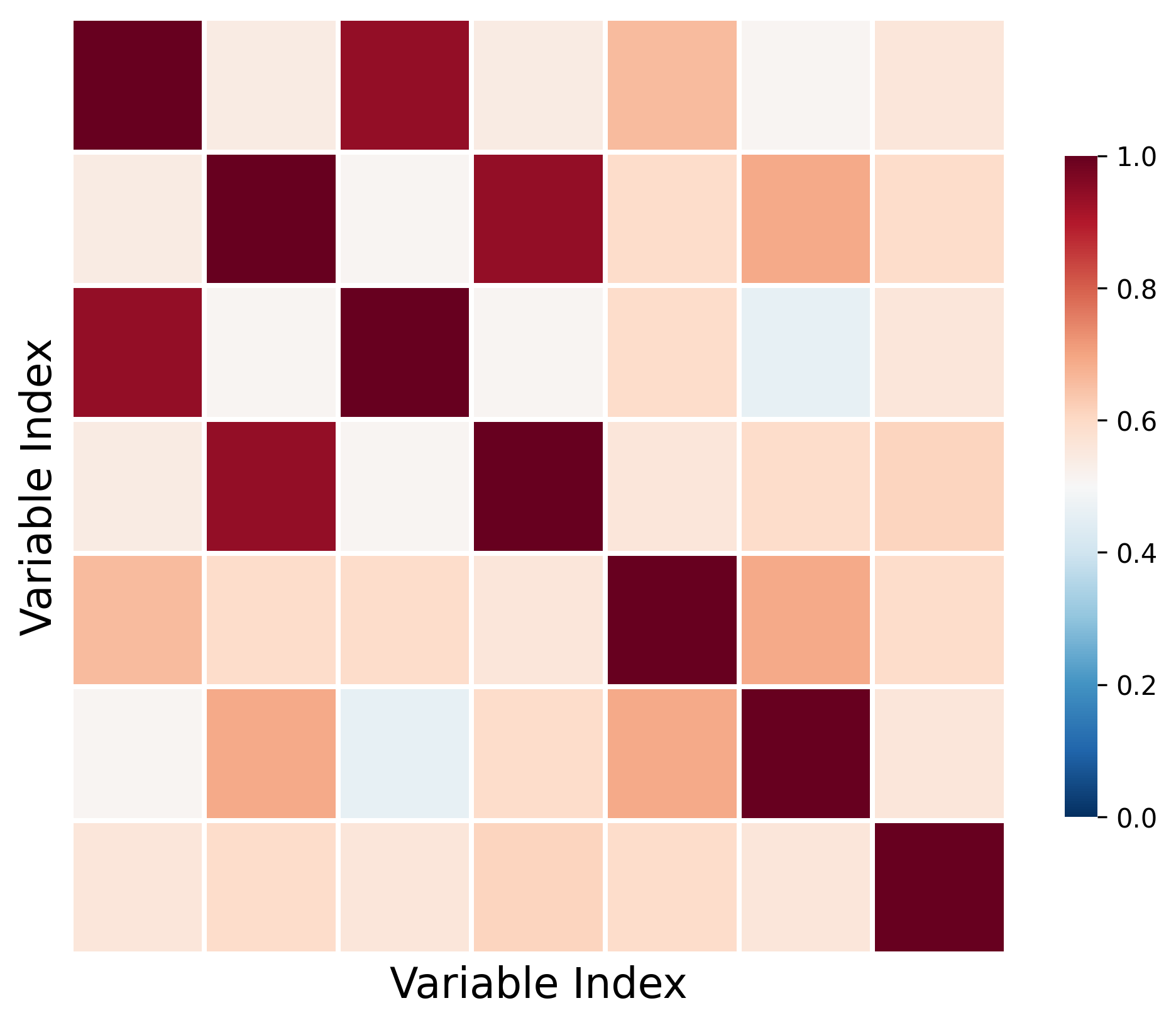}}
\caption{Heatmap of learned variable correlations on the ETTh1 dataset}
\label{fig3}
\end{figure}
We visualize the learned probability matrix $P$ extracted from the ETTh1 dataset. As illustrated in Fig.\ref{fig3}, the heatmap explicitly displays the learned dependency structure among the 7 variables. The TFG module successfully assigns higher relational probabilities to highly correlated variable pairs, while automatically filtering out noisy or irrelevant connections by pushing their weights near zero. This visualization confirms that TFGformer not only achieves strong forecasting accuracy but also provides interpretable insights into the underlying variable interactions.

\subsection{Sensitivity Analysis}
% As shown in Table \ref{tab:sensitivity}, the model achieves the best performance (lowest MSE) on the Weather dataset when $\tau=0.5$. Performance degrades when the temperature is too low (e.g., $\tau=0.1$), which may lead to an overly sparse graph filtering out useful dependencies. Conversely, higher temperatures (e.g., $\tau \ge 1.0$) result in a denser graph, reintroducing noise from irrelevant variables. This confirms that our TFG module effectively benefits from learning a balanced, sparse dependency graph. While $\tau=0.5$ serves as a robust default, the optimal $\tau$ varies slightly across datasets due to differing variable dimensions and inherent sparsities. Therefore, in practice, $\tau$ is treated as a tunable hyperparameter optimized on the validation set for each specific dataset.
Table \ref{tab:sensitivity} shows the lowest MSE on the Weather dataset is achieved at $\tau=0.5$. Performance degrades at extreme temperatures: lower values (e.g., $\tau=0.1$) over-sparsify the graph and filter out useful dependencies, while higher values ($\tau \ge 1.0$) reintroduce noise. This confirms the necessity of learning a balanced, sparse graph. Although $\tau=0.5$ is a robust default, the optimal $\tau$ varies with dataset characteristics (e.g., dimensionality) and is thus treated as a tunable hyperparameter in practice.
\begin{table}[h]
\centering
\caption{Sensitivity analysis of Gumbel-Softmax temperature ($\tau$) on the Weather dataset (Avg. MSE).}
\label{tab:sensitivity}
\setlength{\tabcolsep}{8pt} % 调整列间距
\begin{tabular}{l|ccccc}
\toprule
\textbf{Temperature ($\tau$)} & 0.1 & 0.5 & 1.0 & 2.0 & 5.0 \\
\midrule
\textbf{MSE} & 0.173 & \textbf{0.162} & 0.168 & 0.182 & 0.187 \\
\textbf{MAE} & 0.211 & \textbf{0.209} & 0.224 & 0.231 & 0.232\\
\bottomrule
\end{tabular}
\end{table}

\section{Conclusion}
% We propose TFGformer, a novel framework for multivariate time series forecasting. It features a Time-Frequency Graph (TFG) module—using STFT and a learnable Mahalanobis distance to build sparse, dynamic dependency graphs that guide attention and filter noise—alongside an MLP-based Covariate Fusion Module (CFM) to integrate historical and future contexts. Experiments on seven benchmarks demonstrate TFGformer achieves state-of-the-art performance on six datasets. Future work will extend this architecture to anomaly detection, imputation, and large-scale foundation models.
We propose TFGformer, a novel framework for multivariate time series forecasting. It features a Time-Frequency Graph module that uses STFT and a learnable Mahalanobis distance to build sparse, dynamic dependency graphs for guiding attention and filtering noise, alongside an MLP-based Covariate Fusion Module to integrate historical and future contexts. Experiments on seven benchmarks demonstrate TFGformer achieves superior performance on six datasets. Future work will extend this architecture to anomaly detection, imputation, and large-scale foundation models.

\end{document}